\documentclass[10pt,twocolumn,letterpaper]{article}

\usepackage{wacv}              

\usepackage{graphicx}

\usepackage{amsmath}
\usepackage{amssymb}
\usepackage{booktabs}
\usepackage{times}
\usepackage{epsfig}
\usepackage{multirow}
\usepackage{siunitx}
\usepackage{tabularx}
\usepackage{tikz}
\usepackage{pgfplots}
\usepackage{pgfplotstable}
\usepackage{caption}
\usepackage{enumitem}
\usepackage[accsupp]{axessibility} 
\usepackage[pagebackref,breaklinks,colorlinks]{hyperref}

\usepackage[capitalize]{cleveref}
\crefname{section}{Sec.}{Secs.}
\Crefname{section}{Section}{Sections}
\Crefname{table}{Table}{Tables}
\crefname{table}{Tab.}{Tabs.}

\def\wacvPaperID{1993} 
\def\confName{WACV}
\def\confYear{2024}

\begin{document}

\title{Self-Supervised Learning for Visual Relationship Detection through Masked Bounding Box Reconstruction}


\author{Zacharias Anastasakis $^{1,4}$\\
{\tt\small zaxarisanastasakis@gmail.com}
\and
Dimitrios Mallis $^{2}$\\
{\tt\small dimitrios.mallis@uni.lu}
\and
Markos Diomataris $^{3}$\\
{\tt\small mdiomataris@student.ethz.ch}
\and
George Alexandridis $^{4}$\\
{\tt\small gealexandri@islab.ntua.gr}
\and
Stefanos Kollias $^{4}$\\
{\tt\small stefanos@cs.ntua.gr}
\and
Vassilis Pitsikalis $^{1}$\\
{\tt\small vpitsik@deeplab.ai}
\vspace{0.1cm}
\and
$^{1}$ Deeplab, Athens
\and
$^{2}$ SnT, University of Luxembourg
\and 
$^{3}$ ETH, Zürich 
\and
$^{4}$ National Technical University of Athens
}
\maketitle

\begin{abstract}
We present a novel self-supervised approach for representation learning, particularly for the task of Visual Relationship Detection (VRD). Motivated by the effectiveness of Masked Image Modeling (MIM), we propose Masked Bounding Box Reconstruction (MBBR), a variation of MIM where a percentage of the entities/objects within a scene are masked and subsequently reconstructed based on the unmasked objects. The core idea is that, through object-level masked modeling, the network learns context-aware representations that capture the interaction of objects within a scene and thus are highly predictive of visual object relationships. We extensively evaluate learned representations, both qualitatively and quantitatively, in a few-shot setting and demonstrate the efficacy of MBBR for learning robust visual representations, particularly tailored for VRD. The proposed method is able to surpass state-of-the-art VRD methods on the Predicate Detection (PredDet) evaluation setting, using only a few annotated samples. We make our code available at \url{https://github.com/deeplab-ai/SelfSupervisedVRD}.
\end{abstract}

\section{Introduction}
\label{sec:intro}

\begin{figure}[!ht]
\setlength{\belowcaptionskip}{-0.5cm}
\begin{center}
   \includegraphics[width=0.90\linewidth]{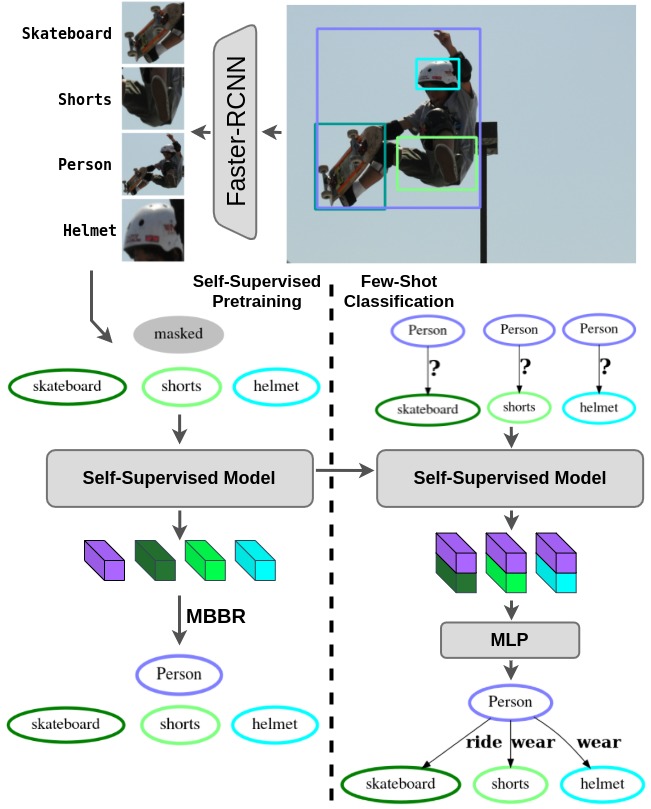}
\end{center}
\vspace{-0.1cm}
   \caption{High-level overview of Masked Bounding Box Reconstruction (MBBR), a novel pretext task for self-supervised VRD. Visual representations for entities within a scene are first extracted through a Faster-RCNN object detector. These visual features are then randomly masked and fed into a transformer encoder. Masked representations are reconstructed, conditioned on the unmasked ones. Learned representations are highly predictive of visual relationships and result in strong Predicate Detection performance in a few-shot classification setting.}
\label{fig:vid_arch}
\end{figure}

Developing machines endowed with the ability to interpret and make decisions based on visual inputs is a critical Computer Vision objective. Research in the field of scene understanding aims to analyze an entire scene or image, in a way similar to that of a human observer. Visual Relationship Detection (VRD) provides an effective approach to scene understanding and constitutes an important component of larger vision pipelines for tasks such as image captioning \cite{img_caption} and visual question answering \cite{vqa2, vqa1, Parelli2023InterpretableVQ}. 

VRD goes beyond identifying/classifying individual components or objects, to extracting the relationships between detected entities. A relationship can be defined as a triplet in the form of $<S, P, O>$, which indicates that the subject $S$ correlates with the object $O$ through the predicate $P$, which is usually selected from a finite set of possible predicate classes. In the example of Fig. \ref{fig:vid_arch}, subject \textsc{[Person]} correlates with object \textsc{[Skateboard]} through the predicate \textsc{[Ride]}. Thus, a scene can be represented in a structured way, \ie, a directed graph where the nodes are the detected entities, while the edges represent the relations among them with a direction from subject to the object. 

The emergence of large datasets providing relationship-level annotations has enabled supervised learning as the dominant approach for visual relationship detection \cite{hl-net, reldn, pum, gps-net, ietrans, login, nodis}. However, labeled data for VRD is difficult to obtain while annotating large-size datasets is a quite time-consuming and expensive procedure. Consider that potential per-image relationships can grow exponentially to the number of objects in the scene, requiring multiple annotations to accurately capture the complex web of possible interactions and associations. Given the apparent cost of data annotation, existing datasets only capture the relationships between a few object categories mostly centered around the human-sensing domain, thus limiting a wider range of potential applications on novel object categories. Finally, defining relationships through distinct predicate classes can introduce various ambiguities (multiple predicate classes can be representative of a visual relationship), human annotators are heavily biased and predicate classes commonly demonstrate a long tail distribution on existing datasets \cite{vg200}. Given the identified limitations of the supervised approach, this work focused on the self-supervised learning (SSL) paradigm for visual relationship detection. SSL involves training models using a pretext task where supervision is provided from the data without requiring any manual annotations. Despite impressive performance in both natural language processing \cite{bert,gpt} and various computer vision tasks \cite{vlbert, visual-bert, landm, vit, beit}, self-supervised approaches for VRD have attracted less attention. 

Our proposed method comprises of a 2-stage pipeline. Initially, a model is pre-trained in a self-supervised manner on large-scale datasets and is able to learn meaningful representations which are later fine-tuned and demonstrate strong performance in few-shot relationship classification.
At the core of our architecture lies a transformer encoder that learns to capture object relationships without requiring any manually annotated labels. The transformer is fed object-level visual features, extracted from bounding boxes using Faster-RCNN \cite{faster-rcnn}. We propose \textit{Masked Bounding Box Reconstruction} (MBBR), a pretext task particularly for VRD, where  a percentage of the input features is randomly masked, and the model is charged with predicting the masked representations solely based on the context provided by the unmasked object features. The key insight is that through MBBR, the model learns representations that encapsulate the complex relationships between objects. These context-aware representations, garnered from the interplay of objects within a scene, prove to be highly predictive of predicates, thereby enabling effective VRD (framework overview in Fig. \ref{fig:vid_arch}).

The performance of learned representations is evaluated on few-shot classification. A VRD classifier is trained for a $k$-shot setting, on top of the features derived from our trained transformer encoder. We show that a simple 2-layer Multi-Layer Perceptron (MLP) can achieve state-of-the-art results for few-shot VRD, thus demonstrating the effectiveness of the proposed SSL framework. 

In summary, our contributions are: \textbf{(1)} We propose MBBR, a novel self-supervised pretext task particularly tailored for self-supervised VRD pre-training, that does not require manual relationship-level annotation. \textbf{(2)} Our proposed method can learn rich relationship-aware object representations that can be used for downstream predicate detection in a few-shot setting. \textbf{(3)} We extensively evaluate our proposed approach both quantitatively and qualitatively on both VRD and VG200 datasets, achieving large performance improvements for few-shot Visual Relationship Detection.

\begin{figure*}
	\begin{center}
		\includegraphics[width=0.9\linewidth]{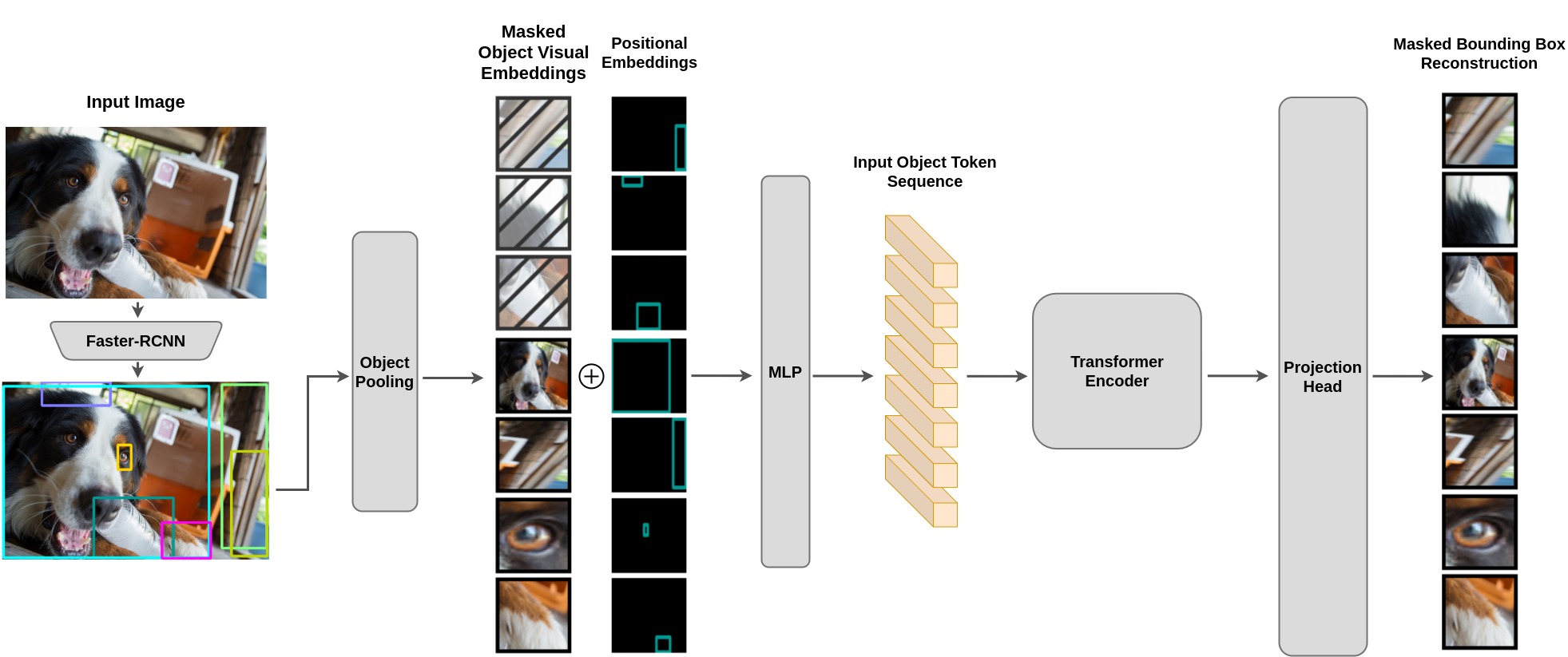}
	\end{center}
	\caption{Overview of the proposed Masked Bounding Box Reconstruction framework. Given an input image, we first extract visual features for each entity in a scene, using a pre-trained Faster-RCNN detector. A percentage of the object embeddings is randomly masked. The network is trained to reconstruct masked objects, from the context provided by unmasked entities.}
	\label{fig:pretraining}
\end{figure*}

\section{Related Work}
\label{sec:related}

\subsection{Visual Relationship Detection}
\label{subsec:related:vrd}

The VRD task was originally introduced in \cite{visual-phrases}, where authors treat each triplet $<S, P, O>$ as an individual class for which a separate classifier is trained. This early formulation required training of $O(N^2K)$ classifiers (where $N$, $K$ is the number of entities and predicates respectively), which was later \cite{language-priors} reduced to $O(NK)$ by training a single classifier per predicate class. Lu \etal
\cite{language-priors} proposed the addition of linguistic priors to exploit entity semantics, leading to enhanced predicate classification. In \cite{neural_motifs}, authors conduct a statistical analysis on the Visual Genome (VG) dataset~\cite{vg200} and discover valuable motifs, \ie, repeated structures and recurring patterns across large scene graphs, which exploit later using bidirectional LSTMs. 

VTransE, proposed in \cite{vtranse}, is a visual translation embedding network which projects the subject $S_{emb}$, object $O_{emb}$ and predicate $P_{emb}$ representations into a low dimensional embedding vector space where \hbox{$S_{emb} + P_{emb} \approx O_{emb}$}. Following \cite{vtranse}, authors in \cite{uvtranse} introduced UVTransE based on the observation that the subtraction of $S_{emb}$ and $O_{emb}$ from the embedding of the union of subject and object, results in the predicate embedding, \ie, \hbox{$U_{emb}- S_{emb} - O_{emb} \approx P_{emb}$}. In \cite{atr} multi-head attention is employed, with a separate head for each predicate class, thus the model is able to attend on multiple visual regions of the input image. Inspired by Faster-RCNN \cite{faster-rcnn} propose Graph-RCNN \cite{graph-rcnn}, a graph network which utilizes a relationship proposal network (RePN) and an attentional graph convolutional network (aGCN), to allow contextual information sharing among objects. Compared to these methods, our proposed approach does not rely on relationship-level supervision for model training. Relationship-aware object representations are learned instead in a self-supervised manner through Masked Bounding Box Reconstruction.


Only a few works explore visual relationship detection under reduced manual supervision. Authors in \cite{energy-based} introduce an energy-based loss for VRD and conduct a few-shot evaluation on relationship triplets. For our few-shot formulation in contrast, a classifier is trained with $k$ samples per predicate class (instead of relationship triplets). To enhance performance under long-tailed distribution for predicate classes of popular VRD datasets \cite{vg200}, authors in \cite{fsl1-predicates-as-functions} propose a two-stage approach. Pre-training is originally performed on the 25 most frequent classes, followed by fine-tuning through few-shot VRD classification on the 25 remaining classes. Predicates are learned as functions, which are then utilized as message-passing mechanisms within a graph convolutional network (GCN). Since \cite{fsl1-predicates-as-functions} focuses on the 25 most frequent predicates, it remains a supervised approach, in contrast to our self-supervised pre-training task that requires no manual annotations. To the best of our knowledge, we are the first to propose a fully self-supervised pre-training pretext task based on bounding box reconstruction, specifically tailored for few-shot Visual Relationship Detection. 

\subsection{Self-Supervised Learning}
\label{subsec:related:ssl}

SSL is the main learning approach used in this work for learning relationship-aware entity representations. It leverages inherent structures and patterns of the data to learn meaningful features \cite{ssl_context_prediction, ssl_image_rotations, ssl_jigsaw, ssl_objects_move}, thus circumventing the need for manual annotations. Several techniques have been proposed in the literature, with contrastive learning and Masked Image Modeling (MIM) recently attracting the most attention. Contrastive methods like SimCLR \cite{simclr} or MoCo \cite{He2019MomentumCF} can learn strong representation by comparing augmented versions of the same image (positive pairs) with distinct images (negative samples). MIM based methods learn through masking a part of an image that is subsequently reconstructed. He \etal \cite{mae} propose to mask a significant portion of an image (75\% of the image patches) and use autoencoders for patch reconstruction. In BEit \cite{beit}, the model is pre-trained on discrete visual tokens from a randomly masked section of the image, that are obtained from the latent representations of discrete variational autoencoders \cite{zeroshot_text_to_image_generation}. Related to ours is the pretext task proposed in \cite{9157808} for image manipulation instead of predicate detection. Compared to our work, authors in \cite{9157808} explore a complex network architecture that also leverages semantic features and perform reconstruction in the pixel space instead of the feature space, utilizing generative networks. In this work, motivated by the recent success of MIM for representation learning based on the transformer architecture, we propose an SSL pipeline based on Masked Bounding Box Reconstruction, particularly for VRD representation learning.

\begin{figure*}
	\begin{center}
		\includegraphics[width=1.0\linewidth]{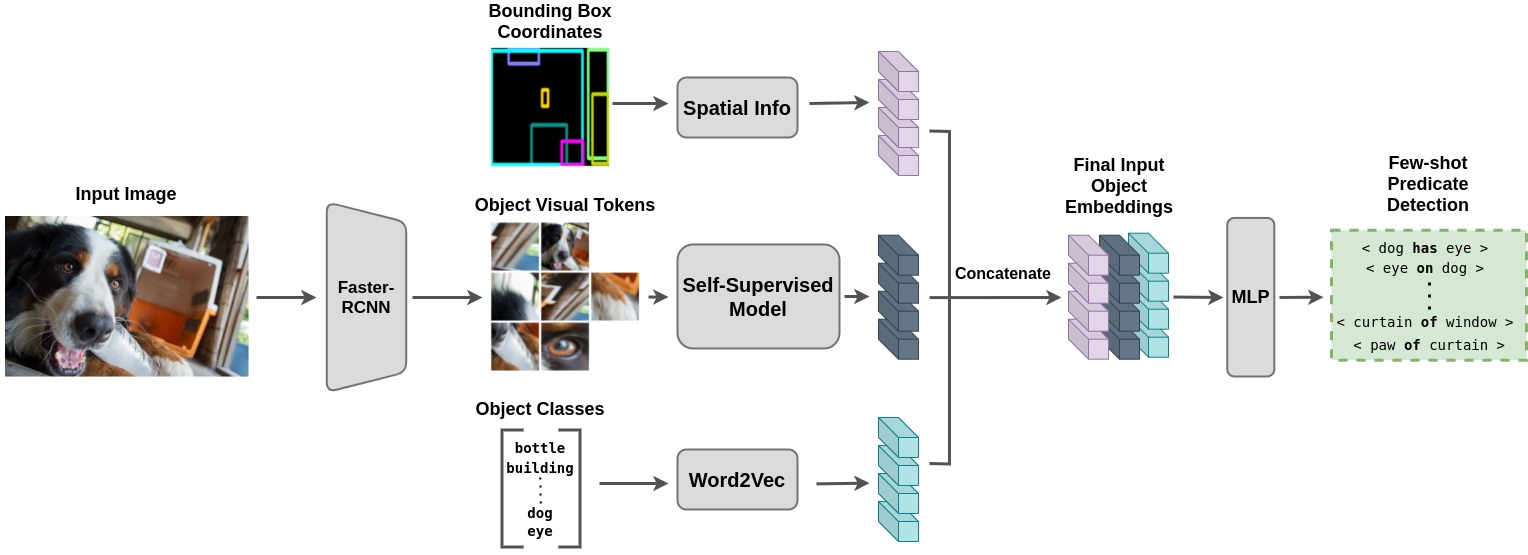}
	\end{center}
	\caption{Architecture of the VRD $k$-shot classifier. Initially, an input image is passed through a pre-trained Faster-RCNN detector, extracting entity-level visual features. Our self-supervised model maps visual features into context-aware representations that capture complex object interactions. These representations along with spatial \cite{atr} and linguistic features (word2vec embeddings~\cite{word2vec} for entity classes) are used to train $k$-shot classifiers for predicate detection.}
	\label{fig:kshot}
\end{figure*}

\section{Approach}
\label{sec:approach}

\subsection{Problem Definition}
\label{subsec:approach:pd}

Let $D=\{I_i, b^S_{\langle i,j \rangle}, b^O_{\langle i,j \rangle}, S_{\langle i,j \rangle}, O_{\langle i,j \rangle}, P_{\langle i,j \rangle}\}$ be a dataset of $N$ images with $i \in [1,N]$, where $b^S_{\langle i,j \rangle}$ and $b^O_{\langle i,j \rangle}$ are the subject / object bounding boxes and $S_{\langle i,j \rangle}$, $O_{\langle i,j \rangle}$, $P_{\langle i,j \rangle}$ denote the subject labels, object labels and predicate labels for the $j^{th}$ relationship in image $I_i$.

In this work, we focus on the Predicate Detection task (PredDet), where the goal is to learn a mapping function 
\begin{equation}
	\Psi :\{ I, b^S, b^O, S, O \} \rightarrow P_k \label{eq:Pk}
\end{equation}
with $k \in [1,K]$ being the set of possible predicate classes. 

We propose to address the above problem through a 2-stage architecture. Firstly (Eq. \ref{eq:z}), a deep neural network with weights $\theta_f$ learns to extract a representation $\boldsymbol{z}$ for each object on a scene in a self-supervised manner without using any object class or predicate class annotations.
\begin{equation}
	\Psi_{\theta_f}: \{ I, b\} \rightarrow \boldsymbol{z}  \label{eq:z}
\end{equation} 
Then in the second stage (Eq. \ref{eq:psi}), a small 2-layer MLP with parameters $\theta_s$ is used to predict 
\begin{equation}
	\Psi_{\theta_s}: \{ I, \boldsymbol{z}^S, \boldsymbol{z}^O, S, O, b^S, b^O\} \rightarrow P_k \label{eq:psi}
\end{equation}
powered by the representations extracted for $\boldsymbol{z}^S$ and $\boldsymbol{z}^O$ during stage-one pre-training. We will describe our proposed architecture for training both $\Psi_{\theta_f}$ and $\Psi_{\theta_s}$ in the following subsection.

\begin{table*}
	\begin{center}
        \small
		\begin{tabular}{|l|cc|cc|}
			\hline
			\multirow{2}{*}{Method} & \multicolumn{2}{c|}{Graph Constraints} & \multicolumn{2}{c|}{No Graph Constraints} \\    
			 & \multicolumn{1}{c}{10-shot} & \multicolumn{1}{c|}{20-shot} & \multicolumn{1}{c}{10-shot} & \multicolumn{1}{c|}{20-shot} \\
			\hline
			Faster-RCNN \cite{faster-rcnn} & $2.6_{\pm 2.81}$ & $14.03_{\pm 5.68}$ & $8.13_{\pm 7.04}$ & $21.30_{\pm 9.48}$  \\
			\hline
			Motifs \cite{neural_motifs} & $2.48_{\pm 3.28}$ & $2.91_{\pm 6.08}$ & $4.51_{\pm 5.05}$ & $3.77_{\pm 6.34}$ \\
			VTransE \cite{vtranse} & $9.75_{\pm 2.55}$ & $14.66_{\pm 3.54}$ & $18.61_{\pm 3.25}$ & $26.98_{\pm 4.07}$  \\
			UVTransE \cite{uvtranse} & $10.41_{\pm 3.29}$ & $15.98_{\pm 2.55}$ & $20.7_{\pm 5.44}$ & $30.67_{\pm 0.88}$  \\
			ATR-Net \cite{atr} & $2.05_{\pm 0.58}$ & $17.88_{\pm 1.94}$ & $6.53_{\pm 1.24}$ & $31.38_{\pm 2.21}$ \\
			\hline \hline
			Our method & $\textbf{20.87}_{\pm 2.46}$ & $\textbf{21.52}_{\pm 1.34}$ & $\textbf{30.75}_{\pm 3.66}$ & $\textbf{34.01}_{\pm 2.51}$  \\
			\hline
		\end{tabular}
	\end{center}
	\caption{Comparison of our method with state-of-the-art on $\{10,20\}$-shot predicate detection on VRD dataset \cite{language-priors}. We report $R@20$ and show the mean value and standard deviation of 5 random initializations.}
	\label{table:vrd_random}
\end{table*}

\begin{table*}
	\begin{center}
        \small
		\begin{tabular}{|l|cc|cc|}
			\hline
			\multirow{2}{*}{Method} & \multicolumn{2}{c|}{Graph Constraints} & \multicolumn{2}{c|}{No Graph Constraints} \\    
			 & \multicolumn{1}{c}{10-shot} & \multicolumn{1}{c|}{20-shot} & \multicolumn{1}{c}{10-shot} & \multicolumn{1}{c|}{20-shot} \\
			\hline
			Faster-RCNN \cite{faster-rcnn} & $1.82_{\pm 1.76}$ & $3.83_{\pm 2.80}$ & $7.91_{\pm 3.80}$ & $10.25_{\pm 6.79}$  \\
			\hline
			Motifs \cite{neural_motifs} & $2.62_{\pm 4.60}$ & $2.39_{\pm 3.73}$ & $6.83_{\pm 3.6}$ & $7.99_{\pm 4.15}$ \\
			VTransE \cite{vtranse} & $7.07_{\pm 3.3}$ & $7.90_{\pm 2.35}$ & $13.70_{\pm 4.18}$ & $17.64_{\pm 4.30}$  \\
			UVTransE \cite{uvtranse} & $5.45_{\pm 2.91}$ & $8.87_{\pm 4.29}$ & $12.74_{\pm 5.5}$ & $20.24_{\pm 8.55}$  \\
			ATR-Net \cite{atr} & $0.52_{\pm 0.43}$ & $3.29_{\pm 4.72}$ & $2.79_{\pm 1.82}$ & $8.46_{\pm 6.73}$ \\
			\hline \hline
			Our method & $\textbf{8.02}_{\pm 1.32}$ & $\textbf{15.37}_{\pm 2.27}$ & $\textbf{16.71}_{\pm 3.76}$ & $\textbf{28.87}_{\pm 2.30}$  \\
			\hline
		\end{tabular}
	\end{center}
	\caption{Comparison of our method with state-of-the-art on $\{10,20\}$-shot predicate detection on VG200 dataset \cite{vg200}. We report $R@20$ and show the mean value and standard deviation of 5 random initializations.}
	\label{table:vg200_random}
\end{table*}

\subsection{Self-supervised representations for VRD}
\label{subsec:approach:ssl}

$\Psi_{\theta_f}$ is trained to reconstruct the feature representation of masked objects, through the context provided by the rest of the objects in the scene (Fig. \ref{fig:pretraining}). Thus, learned features are highly predictive of the relationships between entities and demonstrate robust performance for few-shot predicate detection.

\textbf{Learning Target Formation}: To form the learning target that will be used for MBBR, we initially pass an input image through an off-the-shelf, pre-trained Faster-RCNN to extract visual features. We then apply multi-scale feature pooling for all features inside each entity bounding box $b$, thus extracting a single visual representation $f_{b} \in \mathbb{R}^{256}$ per entity of the scene.

\textbf{Visual geometry embeddings}: Visual geometry embeddings are also used to capture the arrangements of visual embedding within the scene. Following Hu \etal \cite{rel_nets_obj_det}, we represent the position of each image entity by a 4-d vector $\big(\frac{X_{LT}}{W}, \frac{Y_{LT}}{H}, \frac{X_{RB}}{W}, \frac{Y_{RB}}{H}\big)$, where $\big(X_{LT}, Y_{LT}\big)$ and $\big(X_{RB}, Y_{BR}\big)$ are the coordinates of the top left and right bottom corners of the bounding box, respectively, of the entity and $W$, $H$ are the width and height of the input image. This vector is then projected into a high-dimensional space $f_{pos} \in \mathbb{R}^{256}$  by computing sine and cosine functions of different wavelengths.

\textbf{Masked Bounding Box Reconstruction}: We randomly mask each entity feature $f_{b}$ with a probability of $50\%$. Features are then concatenated with geometry embeddings $f_{pos}$ and projected through a linear layer to an entity embedding $f_{e} \in \mathbb{R}^{256}$. For this work, $\Psi_{\theta_f}$ (Eq. \ref{eq:z}) is modelled as a standard feed-forward transformer encoder of \cite{attention_is_all_you_need}. The output of the transformer is a representation per object $z_i$ that is then projected through a linear layer to form the reconstructed entity embeddings $y_{rec,i}$. Our model is trained through a mean square error (MSE) loss between reconstructed entity embeddings and the input embeddings $f_{b,i}$ from the pre-trained Faster-RCNN, for all $N$ entities in a scene. The embeddings $z_i$ can be used later for training a simple MLP in a few-show setting, as discussed next.

\subsection{Few-shot Classification}
\label{subsec:approach:fs}

After self-supervised pre-training, we keep only the encoder without the projection head, and treat the derived representations $z_i$ as visual features for training a classifier on few-shot VRD (Fig. \ref{fig:kshot}). In addition to the visual features, we employ linguistic features, \ie, the word2Vec \cite{word2vec} embeddings of each entity's label, and spatial features \cite{atr} for modeling the respective location of the subject-object pairs. The few-shot classifier takes as input the concatenation of the above features and is trained for predicate class prediction using a standard cross-entropy loss. 

\begin{figure*}
	\begin{center}
		\includegraphics[width=1.0\linewidth]{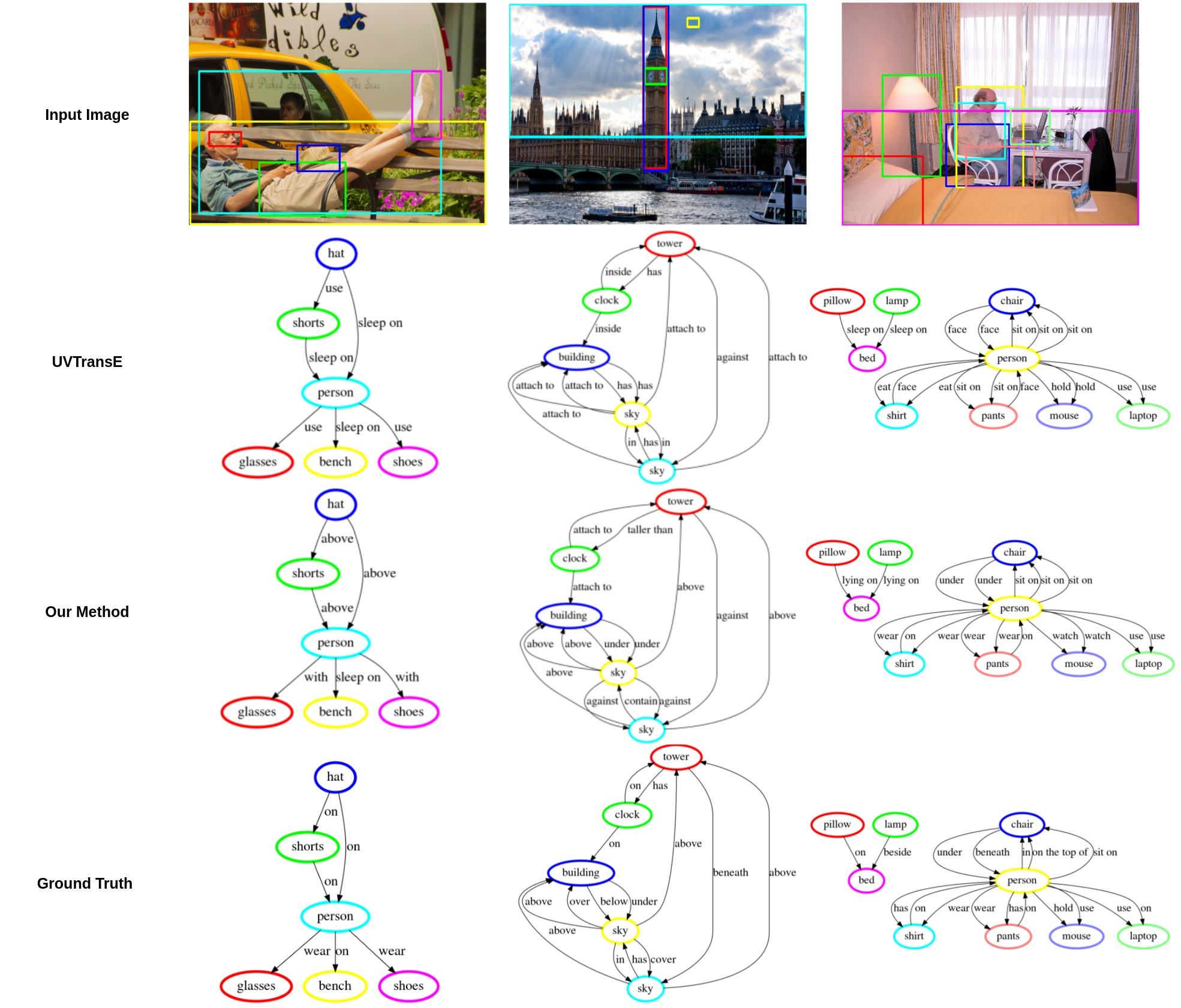}
	\end{center}
	\caption{Directed graphs extracted from 10-shot classifiers on the VRD dataset \cite{language-priors}. We provide a qualitative comparison of our method and UVtranse \cite{uvtranse}. Even in cases where our method does not recover the ground truth triplet, predicted predicates remain semantically accurate. For example see groundtruth relationship $<$ \textsc{building}, \textsc{below}, \textsc{sky} $>$ \textit{(second column)}. Our method recovers a plausible triplet $<$ \textsc{building}, \textsc{under}, \textsc{sky} $>$ compared to $<$ \textsc{building}, \textsc{has}, \textsc{sky} $>$ for UVTransE.}
 
	\label{fig:graphs}
\end{figure*}

\section{Experiments}
\label{sec:experiments}

\subsection{Datasets and Metrics}
\label{subsec:experiments:datasets}

Evaluation is performed on two commonly used publicly available datasets, namely VRD \cite{language-priors} and VG \cite{vg200}. The first is a widely used for visual relationship detection. It contains $5{,}000$ images with 100 object and 70 relationship categories. We use the same split as \cite{language-priors}, \ie, $4{,}000$ training images and $1{,}000$ test images. The total number of annotated triplets is $203{,}284$ in the training set and $7{,}624$ in the test set.

Visual Genome is one of the largest datasets in visual relationship detection. It contains $108{,}077$ images, $3{,}8$ million annotated objects and $2{,}3$ million annotated triplets. We follow the same train/test split as in \cite{vg200}, \ie, $75{,}651$ training images and $32{,}422$ testing images with 150 object classes and 50 relation classes.

In this paper, we focus on the PredDet task. Our proposed SSL pipeline is evaluated in a few-shot setting. Note that in our formulation, $k$-shot refers to $k$ samples per predicate class. Thus, for 10-shot evaluation, in VRD we will use $70 \times 10$ relationships, where $70$ is the number of predicate classes. We use $Recall_k@N \ (R_k@N)$ as our evaluation metric. Given an input image with $x$ subject-object pairs,  $R_k@N$ considers only the top-$k$ predictions for each pair and then selects the $N$ most confident out of a total of $x \cdot k$ predictions. Following \cite{atr}, we refer to evaluation with $k$ = 1 as $graph \ constraints$, indicating that only one edge between entities is allowed. Larger values of $k$ are signified as $no \ graph \ constraints$, allowing multiple edges between entities. In this work, when referring to $no \ graph \ constraints$, $k$ is set to 50 and 70 for the VG200 and VRD, respectively. 


\begin{table*}
	\begin{center}
		\begin{tabular}{|l|ccc|ccc|}
			\hline
			\multirow{2}{*}{Method} & \multicolumn{3}{c|}{Graph Constraints} & \multicolumn{3}{c|}{No Graph Constraints} \\    
			 & \multicolumn{1}{c}{1-shot} & \multicolumn{1}{c}{2-shot} & \multicolumn{1}{c|}{5-shot} & \multicolumn{1}{c}{1-shot} & \multicolumn{1}{c}{2-shot} & \multicolumn{1}{c|}{5-shot} \\
			\hline
			Faster-RCNN \cite{faster-rcnn} & $4.23_{\pm 2.87}$ & $4.24_{\pm 2.4}$ & $5.2_{\pm 5.34}$ & $9.65_{\pm 2.36}$ & $10.52_{\pm 2.12}$ & $12.6_{\pm 4.55}$  \\
			\hline
			Motifs \cite{neural_motifs} & $0.1_{\pm 0.1}$ & $1.15_{\pm 2.35}$ & $9.62_{\pm 7.97}$ & $2.28_{\pm 2.54}$ & $2.36_{\pm 2.55}$ & $11.76_{\pm 8.17}$  \\
			VTransE \cite{vtranse} & $7.46_{\pm 1.55}$ & $9.43_{\pm 2.69}$ & $12.10_{\pm 1.45}$ & $14.43_{\pm 1.05}$ & $16.6_{\pm 3.90}$ & $23.44_{\pm 1.63}$  \\
			UVTransE \cite{uvtranse} & $4.47_{\pm 2.84}$ & $7.52_{\pm 2.88}$ & $12.6_{\pm 0.98}$ & $10.95_{\pm 5.95}$ & $14.88_{\pm 4.59}$ & $23.62_{\pm 1.37}$  \\
			ATR-Net \cite{atr} & $1.43_{\pm 1.30}$ & $1.17_{\pm 0.62}$ & $2.67_{\pm 0.84}$ & $3.96_{\pm 1.92}$ & $5.28_{\pm 2.56}$ & $8.75_{\pm 0.69}$  \\
			\hline \hline
			Our method & $\textbf{9.53}_{\pm 1.88}$ & $\textbf{11.98}_{\pm 2.52}$ & $\textbf{19.90}_{\pm 1.00}$ & $\textbf{16.92}_{\pm 3.14}$ & $\textbf{19.22}_{\pm 3.18}$ & $\textbf{29.92}_{\pm 2.83}$  \\
			\hline
		\end{tabular}
	\end{center}
	\caption{Comparison of our method with state-of-the-art on $\{1,2,5\}$-shot predicate detection on VRD dataset \cite{language-priors}. For this experiment, samples used for few-shot learning are selected manually to ensure effective training. }
	\label{table:vrd_handcrafted}
\end{table*}

\subsection{Implementation details}
\label{subsec:experiments:impl}

The transformer encoder we use is comprised of $8$ attention heads, $6$ layers and a feature dimension of $256$. We use the Adam optimizer \cite{adam_optimizer} with a base learning rate $2 \times 10^{-3}$ and weight decay of $10^{-4}$.
The model is trained for $30$ epochs with batch size $16$. After pre-training, we further fine-tune our model on a $k$-shot setting for $20$ more epochs. For all reported experiments, self-supervised pre-training is performed on VG200 \cite{vg200} as in VRD there are not enough images for our model to learn useful representations. All of our models are implemented in PyTorch.

\subsection{Results}
\label{subsec:experiments:res}

In this section, we analyze the effectiveness of our proposed approach, both quantitatively and qualitatively. Since self-supervised representation learning for predicate detection is a previously unexplored area, comparisons are performed w.r.t. recent supervised methods, trained on a few-shot setting \cite{neural_motifs, vtranse, uvtranse, atr}. We also include a \textit{Faster-RCNN} baseline where instead of using $z_i$ derived from our pre-trained SSL model, we use the object-level visual features that are extracted from the Faster-RCNN. 

Tables \ref{table:vrd_random} and \ref{table:vg200_random} summarize the obtained results on both datasets. Evaluation for PredDet is performed in a $10$ and $20$-shot setting. We observe that through self-supervised pertaining, our encoder can learn robust and generalizable representations that surpass both recent supervised methods and the Faster-RCNN baseline in the few-shot setting, thus demonstrating the effectiveness of MBBR.

Additional few-shot results are provided on Table \ref{table:vrd_handcrafted} for $\{1,2,5\}$-shots. For this evaluation, we opt to manually select the few accurate relationships that are used to train our classifiers. The reason is that relationship tuples in both VRD and VG200 can be highly noisy~\cite{vtranse}, \hbox{\eg, $<$ \textsc{sky}, \textsc{has}, \textsc{sky} $>$} and learned classifiers might fail to generalize when trained on a very small number of noisy examples. As previously mentioned, we find that our pre-trained model surpasses all related methods by a large margin. Interestingly, the $5-$shot setting in Table \ref{table:vrd_handcrafted} results in a similar performance to $20-$shots in Table \ref{table:vrd_random}, thus further demonstrating the importance of selecting accurate relationships for few-shot classification.

\begin{figure}
	\centering
	\resizebox{0.85\linewidth}{!}{
	\begin{tikzpicture}
	  \begin{axis}[
	    title={Top-1 Accuracy on Masked Objects},
	    xlabel={Masking Ratio (\%)},
	    ylabel={Accuracy (\%)},
	    xmin=10, xmax=90,
	    ymin=0.0, ymax=50,
	    xtick={10,25,40, 50,60, 75, 90, 100},
	    ytick={0,10, 20,30, 40, 50, 80},
	    legend style={at={(0.99,0.01)},anchor=south east}, 
	    grid=major,
	    nodes near coords={\textcolor{black}{\pgfmathprintnumber\pgfplotspointmeta}},
	    ytick pos=left,
	    xtick style={draw=none}
	]
	    \addplot[mark=*, color=blue] coordinates {
	        (10,21.13)
	        (25,24.56)
	        (50,34.97)
	        (75,30.18)
	        (90,19.21)
	      };
	  \end{axis}
	\end{tikzpicture}}
	\caption{\textit{Top-1} classification accuracy of reconstructed objects \textit{(masked)}, learned by our self-supervised encoder, for various masking ratio values.}
	\label{fig:ablations_mask_ratio}
\end{figure}
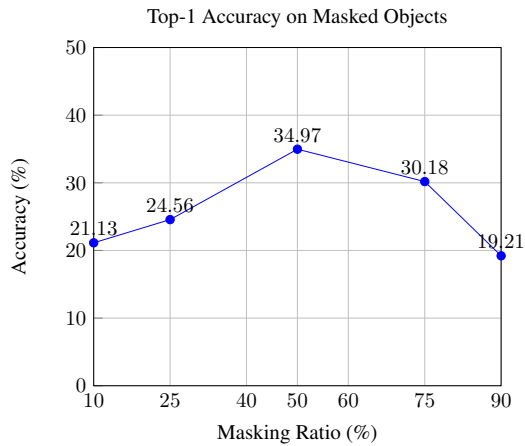

\section{Ablation Studies}
\label{sec:ablation}

\begin{table}
	\begin{center}
        \scriptsize
		\begin{tabular}{|l|cc|cc|}
			\hline
			\multirow{2}{*}{Method} & \multicolumn{2}{c|}{Graph Constraints} & \multicolumn{2}{c|}{No Graph Constraints} \\    
			 & \multicolumn{1}{c}{10-shot} & \multicolumn{1}{c|}{20-shot} & \multicolumn{1}{c}{10-shot} & \multicolumn{1}{c|}{20-shot} \\
                \hline
			\textbf{L + S} & ${13.68}_{\pm 2.27}$ & ${16.02}_{\pm 3.39}$ & ${20.39}_{\pm 1.22}$ & ${26.30}_{\pm 3.87}$  \\
			\textbf{L + S + V} & $\textbf{20.87}_{\pm 2.46}$ & $\textbf{21.52}_{\pm 1.34}$ & $\textbf{30.75}_{\pm 3.66}$ & $\textbf{34.01}_{\pm 2.51}$  \\
			\hline
		\end{tabular}
	\end{center}
	\caption{ Investigation on the impact of linguistic / spatial features. We compare a classifier trained with linguistic / spatial features \textbf{(L}+\textbf{S)} only,  to our full model \textbf{(L}+\textbf{S}+\textbf{V)} utilising self-supervised visual representations \textbf{(V)}. Results for $\{10,20\}$-shot predicate detection on the VRD dataset \cite{language-priors}. We report $R@20$ and show the mean value and standard deviation of 5 random initializations.}
	\label{table:abl_lang_spat}
\end{table}

We conduct an ablation study to further investigate MBBR as a pretext task for VRD representation learning. When evaluating the effectiveness of entity representations learned through unsupervised pretraining, we report \textit{top-1} classification accuracy, for a classifier trained on the reconstructed embeddings $y_{rec,i}$ of masked input objects. Comparison is performed against a baseline classifier trained on Faster-RCNN pooled features $f_{b,i}$.

\begin{figure*}
	\begin{center}
		\includegraphics[width=1.0\linewidth]{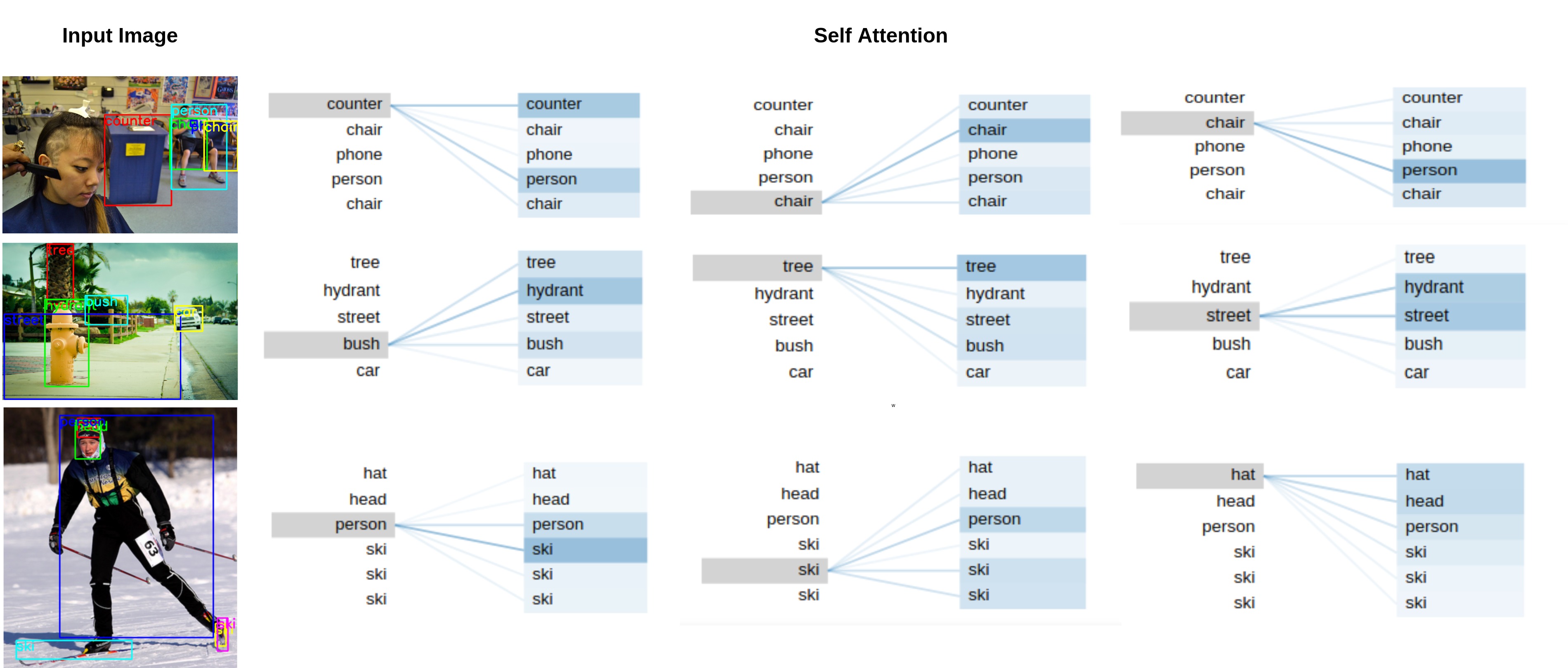}
	\end{center}
	\caption{Visualization of self-attention scores (as in \cite{bertviz}) for our self-supervised transformer model trained with MBBR. We observe that for the effective reconstruction of a specific object, the model attends to other entities with which the reconstructed object is associated through a visual relationship. For instance, the object \textsc{[PERSON]} \textit{(third row)} directs its attection to the object \textsc{[SKI]} with which has a visual relationship, while it shows no attention towards the objects \textsc{[HAT]} or \textsc{[HEAD]} with which it has no association.}
	\label{fig:self-attention-weights}
\end{figure*}

\textbf{Masking Ratio.} We start by investigating the effect of the entity masking ratio used during self-supervised pre-training (Fig. \ref{fig:ablations_mask_ratio}). We find that a masking ratio of $50\%$ results in optimal performance measured in terms of classification accuracy of reconstructed masked objects. Intuitively, a much larger ratio, \ie, $75\%$ or $90\%$ degrades performance since only a few objects remain unmasked and thus provide context for masked object reconstruction. Interestingly, a very small masking ratio of $10\%$ also results in limited performance. Since our encoder is learned through the reconstruction of all object features (not only the masked ones), masking only a small percentage of input objects enables the network to focus on unmasked object reconstruction. Note that as a baseline, the classification of Faster-RCNN representation leads to a $top-1$ accuracy of $74.9\%$ (compared to the $34.9\%$ achieved by our method). Even though our model cannot reach perfect reconstruction of masked objects (compared to an unmasked baseline), we see that learned representations are highly predictive of object relationships and thus achieve strong performance for downstream predicate detection (Tables \ref{table:vrd_random} and \ref{table:vg200_random}).

\textbf{Image vs Feature Masking.} For our proposed method, an image is first passed through a Faster-RCNN to extract input features, a percentage of which is then masked for MBBR. This is in contrast to VLBert \cite{vlbert} where masking is performed on the original input image, \ie, before visual feature extraction. We opt for masking after Faster-RCNN given that for common VRD databases \cite{vg200} large parts of objects on the scene tend to overlap (see qualitative results in Fig. \ref{fig:graphs}). Thus, masking a larger object before feature extraction will also affect the visual features of all overlapping objects.  In our experiments, we observed a $13.63\%$ top- $1$ accuracy, when a $15\%$ masking ratio was applied before Faster-RCNN and $22.78\%$ when it was applied after. Larger masking ratios applied before Faster-RCNN rapidly degrade the effectiveness of MBBR.

\textbf{Reconstruction vs Classification Loss.} Our model is trained to reconstruct inputs through an MSE loss on feature space (similar to the commonly used perceptual loss~\cite{Johnson2016PerceptualLF}), achieving a $R@20$ score of $20.87_{\pm 2.46}$ for $10$-shot predicate detection with graph constraints on the VRD dataset \cite{language-priors} (Table \ref{table:vrd_random} for 5 random initializations). An alternative approach could be learning representations through the prediction of the masked object's class, as in \cite{vlbert}. We find that a classification loss degrades representation learning performance, with a $R@20$ score of $16.7_{\pm 1.51}$ on the same task.

\textbf{Impact of linguistic / spatial features.} For few-shot predicate classification, our method utilises not only visual representations \textbf{V} (learned through MBBR), but also linguistic \textbf{L} and spatial \textbf{S} features (as in \cite{atr}). To quantify the impact of linguistic and spatial features in overall model performance, we compare our complete model (\textbf{L+S+V}) to a variant that only utilises language / spatial features (denoted as \textbf{L+S}). In Table \ref{table:abl_lang_spat}, we show that the addition of visual representation leads to superior performance, with an approximate 8\% increase in $10$-shot and 5.5\% increase in $20$-shot in $R@20$ compared to the variant that does not perceive visual information.

\textbf{Self-Attention Scores.} Through MBBR we learn object representations that are highly predictive of object relationships. To gain further insight into the entity relationships discovered by our self-supervised approach, we provide self-attention visualizations. In Fig. \ref{fig:self-attention-weights} we observe entities tend to attend to other subjects/objects in the scene on which they are connected with a visual relationship. For example in row 3 of Fig. \ref{fig:self-attention-weights}, the \textsc{[ski]}'s attend more to the \textsc{[person]} and other \textsc{[ski]}'s for representation prediction and less to the person's \textsc{[head]} or \textsc{[hat]}.

\textbf{Qualitative Results.} We also include a qualitative comparison with the UVTransE method of \cite{uvtranse} of Fig. \ref{fig:graphs} on $10$-shot setting. Even in cases where our model's predicate prediction differs from  ground-truth, we can see that the detected predicate is semantically more accurate compared to the corresponding predictions of UVTransE. For example see $<$ \textsc{hat}, \textsc{sleep on}, \textsc{person} $>$ for UVTransE compared to our prediction of $<$ \textsc{hat}, \textsc{above}, \textsc{person} $>$ for a ground-truth relationship of $<$ \textsc{hat}, \textsc{on}, \textsc{person} $>$. Overall, our proposed method can produce accurate predicate predictions even when fine-tuned with a very small number of relationship samples.

\section{Conclusion}
\label{sec:conclusion}

We have introduced a self-supervised learning method for visual relationship detection based on Masked Bounding Box Reconstruction (MBBR). MBBR efficiently learns rich context-aware representations that are highly predictive of object visual relationships. Our experiments demonstrate that learned representations can surpass existing methods for downstream VRD in a few-shot setting. 

\vspace{0.1cm}
\noindent
\textbf{Acknowledgements} This project has been funded by \texttt{deeplab.ai}, as part of its research activities, \ie, funding of student research-training and, collaborations with academic institutions. Work was conducted in part, while Z.Anastasakis was an intern with \texttt{deeplab.ai}. 

{\small
\bibliographystyle{ieee_fullname}
\bibliography{submission}
}

\end{document}